\documentclass{article}
\usepackage{spconf,amsmath,graphicx,hyperref,amssymb,bbm,booktabs,pifont,xcolor}
\usepackage{algorithm}
\usepackage{algorithmic}
\usepackage{multirow}
\usepackage{kotex}
\usepackage{enumitem}

\title{AROMMA: Unifying Olfactory Embeddings for Single Molecules and Mixtures}

\name{Dayoung Kang$^1$,\quad JongWon Kim$^2$,\quad Jiho Park$^1$,\quad Keonseock Lee$^2$,\quad Ji-Woong Choi$^{1,2}$,\quad Jinhyun So$^{1,2}$}
\address{
    $^1$Department of Electrical Engineering and Computer Science, DGIST \\
    $^2$Artificial Intelligence Major in Department of Interdisciplinary Studies, DGIST
}

\begin{document}
%
\maketitle
\begin{abstract}
Public olfaction datasets are small and fragmented across single molecules and mixtures, limiting learning of generalizable odor representations.
Recent works either learn single-molecule embeddings or address mixtures via similarity or pairwise label prediction, leaving representations separate and unaligned.
In this work, we propose {\em AROMMA}, a framework that learns a unified embedding space for single molecules and two-molecule mixtures. 
Each molecule is encoded by a chemical foundation model and the mixtures are composed by an attention-based aggregator, ensuring both permutation invariance and asymmetric molecular interactions.
We further align odor descriptor sets using knowledge distillation and class-aware pseudo-labeling to enrich missing mixture annotations.
{\em AROMMA} achieves state-of-the-art performance in both single-molecule and molecule-pair datasets, with up to 19.1\% AUROC improvement, demonstrating a robust generalization in two domains.
\end{abstract}

\begin{figure}[htb]
  \centering
  \includegraphics[width=8.5cm]{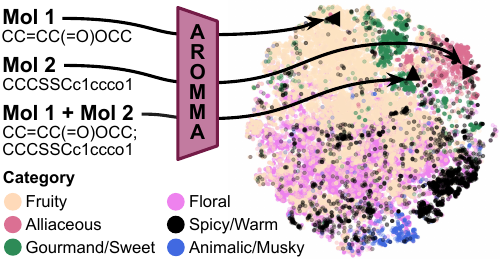}
  \vspace{-0.3cm}
  \caption{Unified embedding space of single molecules and molecule pairs. Each point corresponds to a sample (single molecule or molecule pair), represented by its global embedding from our proposed framework, {\em AROMMA}.}
  \label{fig:embedding_space}
\end{figure}
\vspace{-0.5cm}

\section{Introduction}

\label{sec:intro}
Mapping molecular structure to human olfactory perception remains a fundamental open problem in AI for science. 
Recent progress on single-molecule modeling, exemplified by the Principal Odor Map (POM), shows that a graph neural network (GNN) learned from 5K curated odorants in GoodScents-LeffingWell (GS-LF) datasets, can reach human-level single-molecule description, establishing a strong baseline for structure to odor prediction~\cite{lee2023principal}.
However, POM is limited to single molecules, whereas real-world olfaction involves mixtures with nonlinear interactions~\cite{xu2023odor, barwich2022more, sell2006unpredictability}.

To address this gap, two recent lines of work extend beyond single molecules. 
First, mixture similarity approach proposed in POMMix~\cite{tom2025molecules} aggregates single-molecule embeddings with self-attention and permutation-invariant pooling to learn perceptual distances at the mixture level.
However, it provides no descriptor-level predictions and does not align single and mixture representations in a shared space. 
Second, Sisson et al. train an MPNN-GNN to predict odor descriptors for two-molecule blends by using the Blended Pair (BP) dataset~\cite{sisson2025deep}. 
While an important step toward mixture label prediction, the setup remains non-integrated: its label set is narrower, annotations are sparse, and there is no principled bridge to single-molecule embeddings or a unified odor space. 

In this work, our objective is to learn a unified representation in which single molecules and two-molecule mixtures coexist, enabling descriptor prediction for both and bidirectional knowledge transfer between two domains as depicted in Fig.~\ref{fig:embedding_space}.
However, achieving such a space requires solving three challenges: {\em (i)} Public datasets (e.g., GS-LF and BP) have no canonical descriptor sets and highly heterogeneous label density, {\em (ii)} Mixture perception reflects non-linear effects (agonism/antagonism), which might not be captured by statistic-only pooling used in previous work~\cite{tom2025molecules}, and {\em (iii)} Odor labeled dataset are scarce, demanding robust molecular features to cover broad chemical space.  

To address these challenges, we introduce a novel framework, named {\em AROMMA} (Aggregated Representations of Olfaction via Molecule and Mixture Alignment), with five key features:
\vspace{-0.1cm}
\begin{enumerate}[leftmargin=0.4cm]
    \item {\em AROMMA} learns a unified embedding space that aligns single molecules and mixtures, which enables bidirectional knowledge transfer across two domains. 
    \vspace{-0.2cm}
    \item {\em AROMMA} leverages a chemical foundation model for olfactory tasks, providing robust molecular representations.
    \vspace{-0.2cm}
    \item {\em AROMMA} introduces an aggregator module that applies molecule-wise self-attention followed by cross-attention with a global learnable query, which is permutation-invariant to the order of component while capturing asymmetric component interactions. 
    \vspace{-0.2cm}
    \item {\em AROMMA} integrates single-molecule and molecule-pair datasets. Heterogeneity in the label distribution of different datasets is mitigated by knowledge distillation (KD) and class-distribution-aware pseudo labeling.
    \vspace{-0.2cm}
    \item On GS–LF (single) and BP (pairs) datasets, {\em AROMMA} achieves state-of-the-art AUROC performance with improvements of +3.2\% and +19.1\%, respectively.
\end{enumerate}

\section{Methodology}
\label{sec:method}
\vspace{-0.2cm} 
\subsection{Dataset}
\label{ssec:data}
\vspace{-0.2cm} 

To learn a unified representation, datasets that cover mixtures with varying numbers of molecules are required. We therefore leverage two publicly available resources GS-LF and BP. GS-LF is the most widely used resource for the single molecule case~\cite{lee2023principal, tom2025molecules, sisson2025deep}, curated by olfactory experts (practicing perfumers) \cite{lf}. It comprises about 5K molecules annotated with 138 odor descriptors, where each descriptor is encoded as a binary label reflecting the presence or absence of the perceptual quality, resulting in a multi-label prediction problem. Recent studies have constructed BP, a dataset of approximately 60K molecular pairs by crawling the GoodScents online database~\cite{gs}. Each pair is annotated with 74 potential blend descriptors when two molecules are mixed. This work paved the way for extending olfactory prediction to the multi molecule domain. However, the source explicitly acknowledges that these annotations are incomplete, intended more as a creative tool than a definitive reference~\cite{gs}. Moreover, while GS-LF provides about 5 descriptors per sample, BP is considerably sparser with only 1.4 descriptors on average. This discrepancy in annotation density highlights a distributional gap that can bias model training, as the two datasets reflect different levels of completeness. Beyond this, the absence of a canonical representation for odor descriptors further complicates the task of merging GS-LF and BP. We detail our approach to this challenge in Section~\ref{ssec:phase}.

\begin{figure}[htb]
  \centering
  \includegraphics[width=8.3cm]{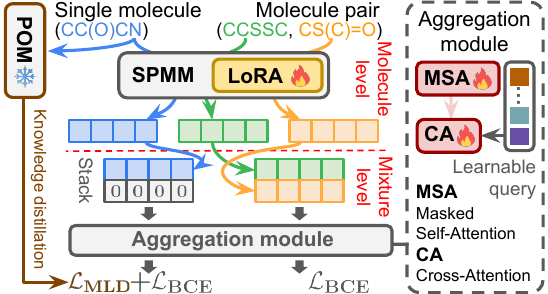}
  \caption{Architecture of {\em AROMMA}. Each molecule is embedded with a chemical foundation model (SPMM)~\cite{chang2024bidirectional} and composed through an aggregation module with a learnable query. Knowledge distillation aligns single-molecule predictions with POM~\cite{lee2023principal}, while the aggregator captures molecular interactions in mixtures.}
  \vspace{-0.2cm} 
  \label{fig:framework}
\end{figure}

\subsection{Overall Architecture}
\label{ssec:architecture}
\vspace{-0.2cm} 
\textbf{Embedder}\quad
Our goal is to enable knowledge transfer between single molecules and molecule pairs, allowing them to share information and ultimately align their representations within a unified embedding space (Fig.~\ref{fig:embedding_space}). Previous studies used the POM architecture for odor prediction or its molecular embeddings for mixture similarity prediction~\cite{tom2025molecules, sisson2025deep}. However, since POM was trained on about 5K single molecules, its limited coverage restricted its ability to generalize to a wider range of chemical compounds. To address this, we employ molecule Structure-Property Multi-Modal foundation model (SPMM)~\cite{chang2024bidirectional} which is pre-trained on about 50M molecules. Whereas other models~\cite{chithrananda2020chemberta, ross2022large} rely exclusively on SMILES (Simplified Molecular Input Line Entry System)~\cite{weininger1988smiles}, a string representation of a molecule, SPMM validates its ability to capture molecular representations by aligning molecular structures with 53 biochemical properties. Given this capacity, we use the SMILES encoder of SPMM as the molecular embedder to obtain robust representations for individual molecules. 
In this process, SPMM is efficiently fine-tuned by using LoRA (low-rank adaptation)~\cite{hu2022lora} with non-stereo SMILES strings provided as input.

\noindent\textbf{Aggregation Module}\quad
In mixtures, certain molecules amplify specific perceptual notes, whereas others may suppress~\cite{sisson2025deep}. Moreover, mixture representation should be permutation invariant, meaning that the model does not depend on the order in which the molecules are presented. To address these aspects, we first apply molecule-wise self-attention, enabling each molecule to interact with others in a mixture \cite{tom2025molecules}. As depicted in Fig.~\ref{fig:framework}, the input to this module is the stacked molecule embeddings $E\in \mathbb{R}^{2\times d_e}$ ($d_e$ denotes the embedding dimension of SPMM), with zero-padding applied for single molecules. These embeddings are linearly projected, passed through a ReLU activation into $E'$, and processed with masked multi-head self-attention to yield $H$, the contextualized molecule embeddings of the input molecules.

While self-attention captures asymmetric molecule-level interactions, permutation invariance must also be preserved at the mixture level. POMMix addressed this requirement by applying Principal Neighborhood Aggregation (PNA), which aggregates molecular information via simple statistics (mean, variance, minimum, maximum) \cite{tom2025molecules}. They also explored cross-attention aggregation, where self-attention outputs served as keys and values and their mean embedding as the query. However, the use of a pre-averaged query inherently compresses molecular information, limiting the capacity to capture fine-grained interactions among components. To address this limitation, we adopt cross-attention with a global learnable query, enabling the model to flexibly capture mixture-level interactions. Similar to the $[\text{CLS}]$ token in BERT, this learnable query serves as a global representation of mixture \cite{devlin2019bert}. 
The self-attention output $H$ is transformed into $H'$, analogous to the projection step in self-attention, which then serves as the keys and values in the multi-head cross-attention with the learnable query $q \in \mathbb{R}^{1 \times d_{h'}}$. We apply LayerNorm so that resulting representation $z$ can serve as the global embedding.

\subsection{Training Phase}
\label{ssec:phase}
\vspace{-0.2cm} 
\begin{figure}[!tb]
  \centering
  \includegraphics[width=8.3cm]{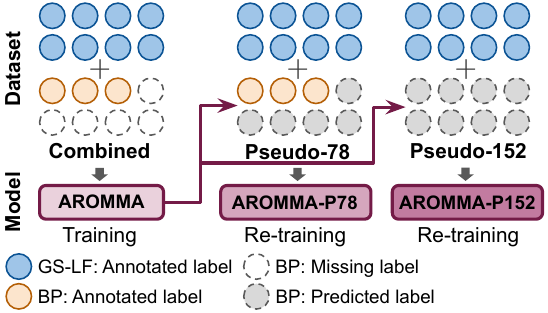}
  \caption{Training strategy of {\em AROMMA}. Missing and sparse labels in the BP dataset are augmented using class-distribution-aware pseudo labeling. The enriched datasets (Pseudo-78/152) are then used to re-train {\em AROMMA}, improving coverage of odor descriptors and performance.}
  \vspace{-0.2cm}     
  \label{fig:phase}
\end{figure}

For training, we first simply integrate the two datasets under a unified descriptor set of 152 labels: GS-LF contributes all 152 labels from its raw data, whereas BP is padded with zeros for the 78 missing labels. As depicted in Fig.\ref{fig:phase}, {\em AROMMA} is initially trained on this unified dataset, and subsequently applied to pseudo-label the missing entries in BP, thereby leveraging unlabeled information~\cite{lee2013pseudo}. Given that the original annotations in BP are potential descriptors, we further re-train the model using pseudo-labels generated across the entire label set. Through this process, we aim to uncover latent labels and enrich the label space. 

\subsubsection{Loss Function and Initial Training}
\label{sssec:learning}
\vspace{-0.2cm} 
As 78 labels in BP are zero-padded, direct training provides no supervision for these labels in molecule pairs, limiting the model’s ability to learn them. 
To address this, we adopt POM as the teacher to match the probability distributions on single-molecule cases. Since our task is formulated as multi-label classification, we adopt the Multi-Label logit Distillation (MLD) loss \cite{yang2023multi}.
\begin{align}
    \mathcal{L}_{\mathrm{MLD}}(P^S, P^T)
    &= \mathcal{D}_{\mathrm{KL}}(P^T \,\Vert\, P^S) \notag \\
    &\quad + \mathcal{D}_{\mathrm{KL}}\big((1-P^T) \,\Vert\, (1-P^S)\big),
\end{align}
where $P^S, P^T$ are softmax outputs of SPMM and POM, and $\mathcal{D}_{\mathrm{KL}}$ denotes Kullback–Leibler (KL) divergence \cite{hinton2015distilling}.
Thus, for single molecules, the objective combines the distillation loss with the Binary Cross-Entropy (BCE) loss:
\begin{equation}
    \mathcal{L}_{\text{BCE}}=-\frac{1}{N}\sum_{i=1}^N[y_i \text{log}(p_i)-(1-y_i) \text{log}(1-p_i)],
\end{equation}
where $N$ is the number of samples, $y_i$ denotes the true label of the $i$-th sample and $p_i$ is the predicted probability. For molecule pairs, distillation is not applied, only the BCE loss is used.
The overall loss function is defined as follows, balancing the two terms equally:
\begin{equation}
\mathcal{L}_{\text{total}} = \alpha \mathcal{L}_{\text{single}} + (1-\alpha)\mathcal{L}_{\text{pair}}, \quad \alpha=0.5,
\end{equation}
where $\mathcal{L}_{\text{single}} = \mathcal{L}_{\text{MLD}} + \mathcal{L}_{\text{BCE}}$ and $\mathcal{L}_{\text{pair}} = \mathcal{L}_{\text{BCE}}$ are losses of singles and pairs datasets, respectively, and $\alpha$ is a loss balancing coefficient. 

\subsubsection{Augmenting Latent Labels and Re-Training}
\label{sssec:pseudo}
\vspace{-0.2cm} 

We initially filled the 78 missing labels in BP with zeros, but these should not be regarded as truly absent but rather as unknown labels. Even the 74 labels that already annotated represent only a subset of the potential odor descriptors. This means that beyond the annotated labels, molecule pairs may still possess unobserved additional descriptors. This motivates us to employ pseudo labeling to uncover such latent labels and enrich the descriptor sets.
Pseudo-labeling is a widely used technique to leverage unlabeled data in semi-supervised learning \cite{lee2013pseudo}. By incorporating pseudo-labeled samples into training, we aim to re-train our model and thereby expand the coverage of the odor descriptor space. We use the class-aware pseudo-labeling method \cite{xie2023class}, which leverages the class distributions observed in the labeled data so that the pseudo-labels better approximate the true label distribution. To ensure consistency across all classes, thresholds are derived from GS-LF, which covers the entire set of 152 labels. In the following equation, $\gamma_c$ denotes the proportion of positive labels for class $c$ in the labeled data.
\begin{equation}
    \gamma_c=\frac{\sum_{j=1}^n \mathbb{I}(y_{jk}=1)}{n}.\\
\end{equation}
The class specific threshold $\tau_c$ is then chosen such that the fraction of predicted probabilities exceeding $\tau_c$ matches $\gamma_c$:
\begin{equation}
    \frac{\sum_{j=1}^n\mathbb{I}(p_{jc} \ge \tau_c)}{n} = \gamma_c.
\end{equation}
Following this procedure, we constructed two augmented datasets, Pseudo-78 and Pseudo-152, and re-trained our model on each. We refer to the resulting models as {\em AROMMA-P78} and {\em AROMMA-P152}, respectively.

\section{Results}
\label{sec:results}
\vspace{-0.2cm} 
\begin{table}[!t]
\centering
\resizebox{0.48\textwidth}{!}{%
\begin{tabular}{lcccccc}
\toprule
Model & Combined & GS-LF~\cite{lee2023principal} & BP~\cite{sisson2025deep} \\
\midrule 
POM \cite{tom2025molecules}
& - & 0.874±0.001 & - \\
MPNN-GNN \cite{sisson2025deep}
& - & 0.851±0.003 & 0.734±0.019\\
AROMMA (ours) & \textbf{0.939±0.002} & 0.900±0.001 & 0.861±0.006 \\
AROMMA-P78 (ours) & 0.931±0.003 &\textbf{0.902±0.001}&0.859±0.008\\
AROMMA-P152 (ours)& 0.932±0.004&\textbf{0.902±0.001}&\textbf{0.874±0.009}\\
\bottomrule
\end{tabular}}
\vspace{-0.2cm}
\caption{Average AUROC($\uparrow$) scores on GS-LF (single molecules), BP (mixtures), and their combination.}
\label{tab:performance}
\end{table}

\noindent\textbf{Implementation}\quad We implement three variants of {\em AROMMA} as explained in Section~\ref{ssec:architecture}. All variants are trained with 4 attention heads. SPMM embeddings ($d_e$=768) are first projected to $d_e=196$ for self-attention, and the outputs are further expanded to $d_{h'}=384$ (equal to $d_q$). For LoRA, we set the rank $r=4$ and the scaling factor $\alpha=8$, resulting in only 110K (0.26\%) trainable parameters. Training is performed with Adam \cite{kingma2014adam} ($\text{lr}=4\text{e-}5$) and early stopping ($\text{patience}=20$). The code and unified datasets are available at https://github.com/DGIST-Distributed-AI-Lab/aromma.

\noindent\textbf{Evaluation}\quad
We evaluate our approach on GS-LF (138 labels for single molecules) and BP (74 labels for molecule pairs), using AUROC with macro-averaging as in prior works~\cite{tom2025molecules,sisson2025deep}. For fair comparison, augmented datasets were used only for training, while evaluation was carried out on the combined set shown in Fig.~\ref{fig:phase}, which merges GS-LF and BP into a unified benchmark built solely from public data.
To implement this evaluation, we used stratified 5-fold splits for GS-LF~\cite{lee2023principal} and predefined folds for BP~\cite{sisson2025deep}. For joint training and evaluation, we synchronized the GS-LF and BP folds, with each fold containing on average 42,555 training, 10,874 validation, and 10,960 test samples.
We then compared {\em AROMMA} against state-of-the-art models, POM~\cite{tom2025molecules} and MPNN-GNN~\cite{sisson2025deep}. The results are reported in Table~\ref{tab:performance}.\begingroup
\renewcommand\thefootnote{\fnsymbol{footnote}}
\endgroup Compared to prior methods, our best-performing {\em AROMMA} achieves improvement of 3.2\% on GS-LF and 19.1\% on BP, establishing a new state-of-the-art on both benchmarks.

\noindent\textbf{Effect of Pseudo Labeling}\quad
For class aware pseudo labeling, thresholds followed the positive label ratios from GS-LF. Interestingly, the average number of active labels per sample in BP also became similar to GS-LF after pseudo-labeling. On average, GS-LF carry about five labels, while BP were originally annotated with only 1.4, rose to 2.7 in Pseudo-78 and 5.6 in Pseudo-152. This adjustment narrows the gap in label density between BP and GS-LF, effectively recovering latent annotations and alleviating the sparsity of BP labels, thereby providing richer supervision for model training. As shown in Table~\ref{tab:performance}, pseudo-labeling was applied only to BP, yet it also leads to performance gains on GS-LF. The results suggest that the pseudo-labels effectively captured the underlying odor manifestation patterns, enabling knowledge transfer from mixtures to single molecules.

Beyond this, pseudo-labeling enriched the semantic granularity. Whereas BP provides broad categories such as floral, GS-LF includes diverse subtypes (e.g., lavender, jasmine, muguet). For instance, when OC1COC(Cc2ccccc2)OC1 and OCc1ccccc1 were blended, BP annotated the pair as floral and fruity. Pseudo-labeling further uncovered additional notes from the floral family, notably rose, thereby capturing latent but unrepresented in the original BP annotations.

\noindent\textbf{Ablation Study}\quad
\begin{table}[!t]
\centering
\resizebox{0.48\textwidth}{!}{%
\begin{tabular}{lcccccc}
\toprule
Embedder & Agg & KD & Combined & GS-LF~\cite{lee2023principal} & BP~\cite{sisson2025deep} \\
\midrule 
POM(freeze)&CA&&0.898±0.003&0.817±0.006&0.851±0.010\\
SPMM(freeze)&CA&\checkmark & 0.935±0.003 & 0.897±0.001 & 0.853±0.007 \\
SPMM(LoRA)&PNA&\checkmark & 0.859±0.006 & 0.881±0.003 & 0.848±0.009 \\
SPMM(LoRA)&CA& & 0.890±0.004 & 0.769±0.005 & 0.830±0.012 \\
SPMM(LoRA) (ours)&CA&\checkmark&\textbf{0.939±0.002}&\textbf{0.900±0.001}&\textbf{0.861±0.006}\\
\bottomrule
\end{tabular}}
\vspace{-0.2cm}
\caption{Ablation study on the main components of {\em AROMMA}: embedder (POM vs.\ SPMM), aggregator (cross-attention (CA) vs.\ PNA), knowledge distillation (KD), and LoRA.}
\label{tab:ablation}
\end{table}
We empirically validate the main components of {\em AROMMA} as summarized in Table~\ref{tab:ablation}. First, when using POM as the embedder, the model generates odor-relevant embeddings, while its performance remains inferior to SPMM. As POM serves as the teacher, KD is not applied in this setting. Second, evaluating aggregation strategies indicates that a learnable query better captures mixture interactions than statistic-based pooling. Third, KD enriches single-molecule predictions, which transfer to mixtures and improve their performance. Finally, LoRA reveals that mixture prediction benefits from updating the embedder, going beyond frozen single-molecule representation.

\vspace{-0.2cm} 
\section{Conclusion}
\label{sec:conclusion}
\vspace{-0.2cm} 
In this work, we proposed {\em AROMMA}, a unified framework that bridges single molecules and mixture of two molecules within a shared embedding space. By leveraging chemical foundation model, attention-based aggregator, and knowledge transfer through distillation and pseudo-labeling, {\em AROMMA} effectively captures asymmetric molecular interactions while alleviating label sparsity. 
Our experiments demonstrate state-of-the-art performance on both GS-LF and BP datasets, with significant gains in the odor prediction of mixtures.
Despite these advances, our study is limited to molecule pairs; however, the proposed architecture can be naturally extended to handle mixtures of three or more molecules, which remains an important future direction.
Also, while {\em AROMMA} focus on non-stereo SMILES representations to ensure scalability, recent advances highlight the potential of 3D molecular representations for richer structural modeling~\cite{zhang2025leveraging}. Incorporating 3D-aware representations into the framework is a promising avenue for future research. 
In this context, {\em AROMMA} highlights a scalable path toward generalizable olfactory AI, opening opportunities for modeling more complex mixtures and enabling future applications in sensory perception and molecular design.

\vfill\pagebreak
\section{Acknowledgment}
\label{sec:acknowledgment}
This work was supported by the Institute of Information \& Communications Technology Planning \& Evaluation (IITP) grant funded by the Korea government (MSIT) (No.RS-2025-02219277, Al Star Fellowship Support (DGIST)), the InnoCORE program of the Ministry of Science and ICT (26-InnoCORE-01), and the Basic Science Research Program through the National Research Foundation of Korea (NRF) funded by the Ministry of Education (RS-2025-25396400).

\bibliographystyle{IEEEbib}
\bibliography{refs}

\end{document}